\documentclass[manuscript]{acmart}

\AtBeginDocument{%
  \providecommand\BibTeX{{%
    \normalfont B\kern-0.5em{\scshape i\kern-0.25em b}\kern-0.8em\TeX}}}

\setcopyright{acmcopyright}
\copyrightyear{2022}
\acmYear{2022}

\usepackage{amsmath}
\DeclareMathOperator*{\argmax}{arg\,max}

\usepackage{multirow}
\newcommand{\blue}[1]{{#1}}

\begin{document}

\title{Modeling Techniques for Machine Learning Fairness: A Survey}


\author{Mingyang Wan}
\email{w1996@tamu.edu}
\affiliation{%
  \institution{Department of Computer Science and Engineering, Texas A\&M University}
  \country{USA}
}
\author{Daochen Zha}
\email{daochen.zha@rice.edu}
\affiliation{%
  \institution{Department of Computer Science, Rice University}
  \country{USA}
}
\author{Ninghao Liu}
\email{ninghao.liu@uga.edu}
\affiliation{%
  \institution{Department of Computer Science, University of Georgia}
  \country{USA}
}
\author{Na Zou}
\email{nzou1@tamu.edu}
\affiliation{%
  \institution{Department of Engineering Technology and Industrial Distribution, Texas A\&M University}
  \country{USA}
}


\begin{abstract}
Machine learning models are becoming pervasive in high-stakes applications. Despite their clear benefits in terms of performance, the models could show \blue{discrimination} against minority groups and result in fairness issues in a decision-making process, leading to severe negative impacts on the individuals and the society. In recent years, various techniques have been developed to mitigate the \blue{unfairness} for machine learning models. Among them, in-processing methods have drawn increasing attention from the community, where fairness is directly taken into consideration during model design to induce intrinsically fair models and fundamentally mitigate fairness issues in outputs and representations. In this survey, we review the current progress of in-processing \blue{fairness} mitigation techniques. Based on where the fairness is achieved in the model, we categorize them into explicit and implicit methods, where the former directly incorporates fairness metrics in training objectives, and the latter focuses on refining latent representation learning. Finally, we conclude the survey with a discussion of the research challenges in this community to motivate future exploration.
\end{abstract}

\ccsdesc[500]{Computing methodologies~Philosophical/theoretical foundations of artificial intelligence}

\keywords{Machine Learning Fairness, Bias Mitigation, Disparate Impact, Disparate Treatment}

\maketitle

\section{Introduction}
\label{sec:1}

Machine learning models are becoming pervasive in real-world applications and have been increasingly deployed in high-stakes decision-making processes, such as loan management~\cite{mukerjee2002multi}, job applications~\cite{raghavan2020mitigating}, and criminal justice~\cite{berk2021fairness}. Despite the clear benefits brought by machine learning techniques, e.g., strong prediction ability by automatically discovering effective patterns in data, the resultant models could show \blue{discrimination} against certain groups of people and lead to fairness issues in practice. For example, it is reported that the deployed automated risk assessment tools for criminal offenders have significant racial disparities: black defendants are almost twice more likely to be falsely flagged as future criminals than white defendants~\cite{angwin2016machine}. The problem is partly caused by the bias that already exists in datasets and could be further amplified by models~\cite{wang2019balanced}, leading to severe negative impacts on the individuals and the society. To mitigate \blue{unfairness} in machine learning models, researchers have developed various techniques at different stages of model development~\cite{kamishima2011fairness,luong2011k,kamiran2012data,madras2018learning,nabi2018fair,pleiss2017fairness,kim2018fairness,noriega2019active}, including pre-processing, in-processing and post-processing methods. Pre-processing~\cite{luong2011k,kamiran2012data,madras2018learning,nabi2018fair,2021} and post-processing~\cite{pleiss2017fairness,kim2018fairness,noriega2019active} are straightforward strategies to \blue{mitigate} \blue{unfairness}: pre-processing methods focus on adjusting the training data distribution to balance the sensitive groups, while post-processing methods calibrate the prediction results after model training.

Unlike the above techniques, in-processing methods directly incorporate fairness into model design, which can induce intrinsically fair models and fundamentally mitigate fairness issues in machine learning models~\cite{hashimoto2018fairness,shen2016disciplined,dwork2012fairness,kearns2018preventing,edwards2015censoring}. First, in-processing techniques can address the problem of bias amplification in model training: the tendency that the model exacerbates biases in the training data. This problem is often caused by the algorithm and cannot be solely attributed to the data~\cite{foulds2020intersectional,wang2021directional}. In-processing methods directly take fairness into consideration in model optimization so that the converged model could \blue{achieve fairness even with biased data as input}~\cite{caton2020fairness}. Second, in-processing methods can effectively fine-tune the representations from pre-trained models to mitigate the bias without huge re-training efforts. While many pre-trained deep learning models~\cite{he2016deep,he2017neural,devlin2018bert} have demonstrated their performance power, they inevitably encode or amplify bias, leading to \blue{potential} unfair consequences in downstream applications. It is with huge efforts and sometimes even infeasible to re-train the models for \blue{unfairness} mitigation. Instead, in-processing methods can be used to fine-tune the representations in pre-trained models. For example, a contrastive learning framework was proposed to debias the representations without re-training for BERT~\cite{cheng2021fairfil}, an indispensable pre-trained component in the modern natural language processing models~\cite{devlin2018bert}. Third, it remains a challenge to develop effective in-processing solutions. The existing algorithms mainly focus on addressing the fairness issue under a single and known sensitive attribute. There exist application scenarios where many of the existing methods will fail. For instance, it is crucial to address fairness issues when sensitive attributes information is not disclosed or available. 


Complementing the previous survey papers that mainly present high-level overview of machine learning fairness and general taxonomies of pre-processing, in-processing, and post-processing methods~\cite{du2020fairness,caton2020fairness,mehrabi2021survey}, we aim to summarize and categorize the key ideas behind the existing in-processing methods for motivating future exploration of algorithmic fairness. In this article, we survey through the in-processing methods for machine learning fairness and categorize them into explicit and implicit mitigation \blue{methods} based on where the fairness is achieved in the model. Explicit mitigation is achieved by formulating the objective function with fairness constraints or regularizers, which are often inspired by the fairness measurements and designed to enforce the predictions to be less dependent on the sensitive attributes. The explicit methods are often flexible and easy to implement as they only require minimum modifications to objective functions. Some example regularizers include co-variance relationships~\cite{kamishima2011fairness}, absolute correlation regularization~\cite{beutel2019putting}, Wasserstein-1 distances~\cite{jiang2020wasserstein}, etc. Alternatively, implicit mitigation focuses on debiasing the representations so that the resulting predictions do not show discrimination towards a minority group or individuals. The implicit approaches often target deep learning models, where learning representations is crucial. For example, they can be used to fine-tune the representations in pre-trained models to debias downstream applications. Some strategies that fall into this category include adversarial learning~\cite{edwards2015censoring}, contrastive learning~\cite{cheng2021fairfil}, disentangled representation learning~\cite{locatello2019fairness}, etc.

The remainder of this article is structured as follows. Section~\ref{sec:2} gives an introduction of the fairness problem and some representative measurements to quantify the (un)fairness. Section~\ref{sec:3} and Section~\ref{sec:4} summarize the existing research on explicit and implicit mitigation of \blue{unfairness}, respectively. \blue{Section~\ref{sec:6} provides general discussions on the differences of the existing in-processing methods.} Section~\ref{sec:6} discusses the research challenges in this community. Finally, we conclude the article in Section~\ref{sec:7}.


\section{Fairness in Machine Learning}
\label{sec:2}
In this section, we introduce the fairness problem in machine learning, measurements for quantifying the degree of bias for different types of fairness, and an overview of the taxonomy for in-processing techniques.

\begin{table}[]
    \centering
    \caption{Main symbols and definitions.}
    \begin{tabular}{l|l} \toprule
    \textbf{Symbol} & \textbf{Definition} \\
    \midrule
    $f_\theta$ & A machine learning model that maps attributes to predictions with parameters $\theta$. \\
    $\textbf{x} \in \mathbb{R}^{d}$ & The attributes with a dimension of $d$.  \\
    $x_s \in \mathbb{R}$ & The sensitive attribute.  \\
    $\textbf{s} $ & A tuple representing all the values of (multiple) sensitive attributes.   \\
    $\hat{y} \in \{-1, 1\}$ & A binary prediction that indicates negative and positive outcomes for $-1$ and $1$, respectively.  \\
    $y \in \{-1, 1\}$ & The ground truth.  \\
    $\mathcal{D}$ & The training dataset.  \\
    $L(\mathcal{D};\theta)$ & The objective function of the machine learning model.  \\
    $\textbf{z} \in\mathbb{R}^{l}$ & The encoded latent representation with a dimension of $l$.  \\
    $R(\mathcal{D};\theta)$ & Fairness regularizer.   \\
    $\Omega(\mathcal{D};\theta)$ & Fairness constraint.   \\
    ${\left \|\theta\right\|}_{2}^{2}$ & $L_2$ regularizer.   \\
    $[\cdot]_+$ & Ignoring the parts less than zero.   \\
    $d(\cdot, \cdot)$ & Distance of two individuals in attribute space.  \\
    $D(\cdot, \cdot)$ & Distance of two individuals in prediction space.  \\
    $\textbf{x}_+$ & A positive sample in contrastive learning   \\
    $\textbf{x}_-$ & A negative sample in contrastive learning   \\
    $\text{MI}(\cdot, \cdot)$ & Mutual information.   \\
    $\textbf{x}_{\bar{s}}$ & The non-sensitive attributes.  \\
    $\textbf{b}$ & The sensitive latents.  \\
    $\hat{x}_{s}$ & The predicted sensitive attribute.  \\
    $A_\phi$ & An adversary network with parameters $\phi$. \\
    $L(\mathcal{D};\phi)$ & The adversarial loss for $\hat{x}_{s}$ and $x_s$. \\
    $L_r(\mathcal{D};\theta)$ & The reconstruction loss. \\

    \bottomrule
    \end{tabular}
    \label{tab:symbols}
\end{table}

\subsection{Fairness Problem} \label{sec:2.1}
We first \blue{discuss fairness in general, and distinguish the concepts of fairness and bias from the psychometrics view. Then, we formally define the fairness problem from the computational perspective, followed by} a supervised binary classification task \emph{Adult} from UCI repository~\cite{asuncion2007uci} as an example to describe the underlying fairness issues.

\blue{\emph{Fairness} is a social and subjective concept describing the appropriateness of how a social construct is measured. Its definition varies across different cultures and societies. In the context of machine learning, unfairness often refers to the situation where a model shows discrimination against certain groups of people (e.g., races and genders). Another term that frequently appears is \emph{bias}, which is often used interchangeably with fairness in the machine learning literature~\cite{mehrabi2021survey}. However, these two concepts are different from the psychometrics view. A recent study has provided a clear and comprehensive discussion to distinguish the two~\cite{booth2021integrating}. Unlike fairness, bias refers to any systematic error that affects the performance of different groups in different ways. Bias is defined as either contamination of the measurement of a construct (e.g., by adding unnecessary gender information) or a deficiency in the measurement (e.g., by not capturing all aspects of the construct which are relevant). While a biased model is likely to be unfair, a biased prediction may not necessarily be unfair. For example, in a hypothetical experiment performed in~\cite{pessach2020algorithmic}, a college admission decision algorithm that is given knowledge about demographics could help admit more people from underrepresented races. The algorithm is inherently biased (while fair) because it uses demographic information. We refer interested readers to~\cite{booth2021integrating} for a more comprehensive discussion.}

\blue{In this article, we use the term \emph{fairness} when an algorithm is designed to achieve good results for specific fairness measurements. For example, explicit methods are often designed to achieve fairness because they directly incorporate fairness measurements into learning objectives. We use the term \emph{bias} when an algorithm can only mitigate the bias but may or may not achieve good fairness results. For example, some implicit methods, such as contrastive learning~\cite{cheng2021fairfil} and disentanglement~\cite{locatello2019fairness}, only seek to remove the bias of sensitive attributes in representations. The fairness performance depends on the downstream tasks and the fairness measurements (detailed in Section~\ref{sec:2.2}) to be used. In this subsection, we use two commonly used measurements, demographic parity, and equal opportunity, as examples to illustrate the underlying fairness issues.}

Without loss of generality, we consider a binary classification problem. Formally, the task is to learn a mapping $f_\theta: \mathbb{R}^{d} \to \mathbb{R}$, which takes as input the attributes $\textbf{x} \in \mathcal{X} \subseteq \mathbb{R}^{d}$ (including a sensitive attribute $x_s \in \mathcal{X}_s \subseteq \mathbb{R}$) for each individual, where $\theta$ is the model weights and $d$ denotes the number of attributes. Typically, we only consider a single sensitive attribute. It is also worth noting that researchers have studied fairness problems where the sensitive attribute information is not fully available~\cite{hashimoto2018fairness}, or there are multiple sensitive attributes~\cite{kearns2018preventing}. In the following discussions, we focus on the most common scenario with a single known sensitive attribute. Given all the attributes, the model will make a binary prediction $\hat{y} \in \{-1, 1\}$ that indicates negative and positive outcomes for $-1$ and $1$, respectively (we use $y \in \{-1, 1\}$ to denote the ground truth). The model is then often trained on a training dataset $\mathcal{D}$ using an objective function $L(\mathcal{D};\theta)$ with the goal that the trained model could make predictions for unlabeled individuals in a hold-out test set. For a deep learning model, the mapping $f_\theta$ usually consists of an encoder that maps $\textbf{x}$ into a latent representation $\textbf{z} \in \mathcal{Z} \subseteq \mathbb{R}^{l}$, where $l$ is the dimension of the latent space, and a decoder that maps $\textbf{z}$ into $\hat{y}$. We summarize the symbols used throughout this article in Table~\ref{tab:symbols}. Regardless of the adopted models, the models could be \blue{unfair in the use of} the sensitive attribute $x_s$. Next, we instantiate the above problem definition with the \emph{Adult} dataset, which will serve as an example in our discussion to facilitate understanding.

The task of the \emph{Adult} dataset is to predict whether a person's salary is higher (a positive outcome) or lower (a negative outcome) than 50K dollars annually. The attributes in \emph{Adult} are tabular, describing the characteristics of each individual. They consist of many non-sensitive attributes, such as occupation, education level, and hours per week, and some sensitive attributes, such as gender and race. We consider a single sensitive attribute $x_s$, which is binary and $x_s \in \{male, female\}$. From the computational perspective, the fairness problem can be generally grouped into two categories: \emph{disparate impact} and \emph{disparate treatment}, which approach the fairness problem from the group- and the individual-level, respectively~\cite{zafar2017fairness2}.

\begin{figure}
    \centering
    \includegraphics[width=\textwidth]{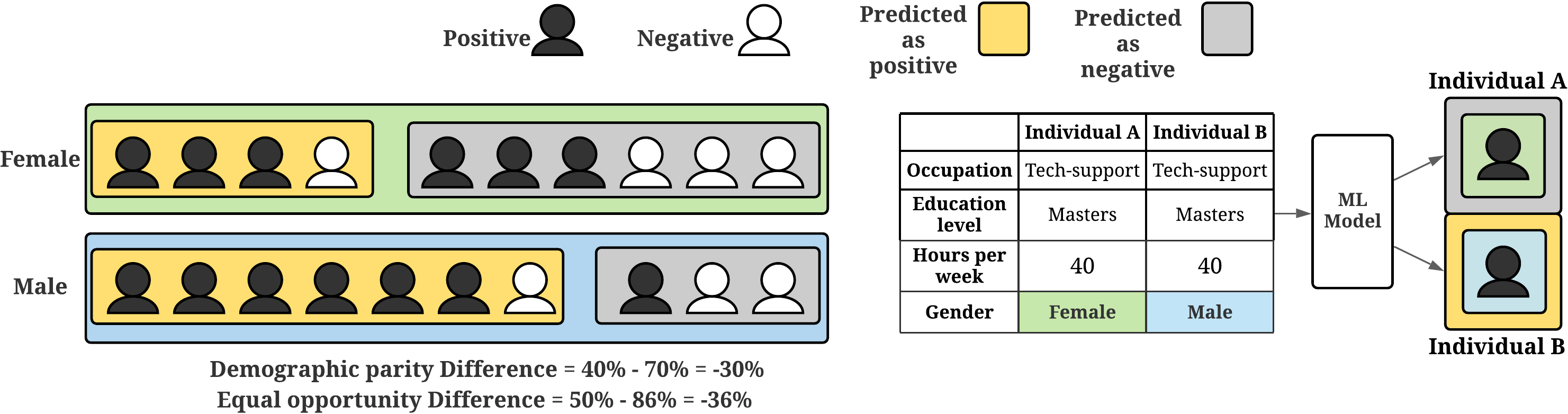}
    \caption{Disparate impact (left) and disparate treatment (right) on the \emph{Adult} dataset, where a binary classifier is trained to predict high salary (positive) or low salary (negative). Disparate impact addresses group-level bias, where the two sensitive groups (male and female) are treated differently according to a fairness metric (e.g., true positive rates are significantly different across groups). Disparate treatment focuses on individual-level bias, where two individuals with different sensitive attributes (i.e., gender) but similar non-sensitive attributes of occupation, education level and hours work per week are treated differently.}
    \label{fig:twobias}
\end{figure}

\subsubsection{Disparate Impact} \hfill\\
Disparate impact refers to the situation where the model disproportionately discriminates certain groups~\cite{zafar2017fairness2}, even if the model does not explicitly leverage the sensitive attribute to make predictions but rather on some proxy attributes. The left-hand side of Figure~\ref{fig:twobias} provides an example of disparate impact in the \emph{Adult} dataset. According to the prediction results, the model is \blue{unfair} because it tends to predict male instances as positive with a higher probability (i.e., 0.7) than females as positive (i.e., 0.4). In the example, the true-positive rate is $50\%$ in the female group while the rate is $86\%$ in the male group. There are various ways and perspectives to measure disparate impact, which will be introduced in details later.

\subsubsection{Disparate Treatment} \hfill\\
Unlike disparate impact that focuses on group-level discrimination, disparate treatment refers to \blue{unfairness} at the level of individuals. The underlying intuition is that a model should not treat individuals with similar attributes differently. The right-hand side of Figure~\ref{fig:twobias} gives an example of disparate treatment in the \emph{Adult} dataset. The two individuals A and B have very similar background information such as the same occupation, equal education level, and identical work hours per week, while gender is the only different attribute between them. In this example, the model is \blue{unfair} because the prediction for individual A is negative while it is positive for individual B. Such a phenomenon usually suggests that the model could have undesirably leveraged sensitive attributes, such as gender information in this example, to make predictions.

\subsection{Fairness Measurements}
\label{sec:2.2}

Choosing the fairness measurement to be applied is crucial when detecting and mitigating model \blue{unfairness}. The chosen measurement decides how fairness is defined, and it reflects the expectation of target applications. Many fairness measurements have been proposed in the literature. This subsection focuses on some representative ones that quantify the disparate impact (group fairness measurements) or the disparate treatment (individual fairness measurements), respectively. Additionally, we introduce a hybrid criteria that measures the fairness of subsets of multiple sensitive attributes. \blue{Table~\ref{tbl:measure} summarizes the measurements according to group/individual measurements, the mechanisms, and whether sensitive attribute is known.} For intuitiveness, we still use the \emph{Adult} dataset as the example to introduce how fairness could be quantified under different measurements.

\blue{Note that, due to the subjectivity of fairness, different fairness measurements could be suitable for different scenarios. We provide more discussions in Section~\ref{sec:5}. In this article, we assume that the appropriate fairness measurements are already chosen in downstream applications, and focus on surveying the design of unfairness mitigation techniques.}

\subsubsection{Group Fairness Measurements} \label{sec:2.2.1} \hfill\\
Group fairness measurements focus on the difference of model predictions on two or more groups. The group that an individual belongs to is indicated by its sensitive attributes. We use $s_i$ and $s_j$ to denote two different sensitive groups associated with a sensitive attribute. For the example task of salary prediction, $s_i$ and $s_j$ represent female and male groups, respectively. In some scenarios, there could be more than more than two sensitive groups. In this case, we can often aggregate the measurement values for each pair of groups as an overall measurement. Following the above notations, we summarize some representative group fairness measurements as below\blue{, where the first measurement is parity-based (i.e., considering positive rates), the next five measurements are based on confusion matrix (i.e., considering aspects of true positive rates, true negative rates, false positive rates, and false negative rates), the next measurement is based on worst-off utility (i.e., the worst value of a metric, such as accuracy, AUC, across groups), and the final one is defined for clustering problems.}

\begin{table}[t]
\footnotesize
\centering
\caption{\blue{A summary of fairness measurements.}}
\label{tbl:measure}
\begin{tabular}{l|l|l|l|l} \toprule
\textbf{Category} & \textbf{Mechanism} & \textbf{Sensitive Attribute} & \textbf{Measurement} & \textbf{Description} \\
\midrule

\multirow{8}{*}{Group} & Parity  & Known & Demographic parity & Equal percentages of positive outcome  \\
\cline{2-5}
~ & \multirow{5}{*}{Confusion Matrix} & \multirow{5}{*}{Known} & Equal opportunity & Equal true positive rate  \\
~ & ~ & ~ & Equal opportunity & Equal true positive rate and false positive rate \\

~ & ~ & ~ & Overall accuracy equality & Equal accuracy  \\
~ & ~ & ~ & Treatment equality & Equal false negative rate / false positive rate \\

~ & ~ & ~ &  Equalizing disincentives & Equal  true positive rate - false positive rate \\
\cline{2-5}
~ & Worst-off utility & Unknown &  Rawlsian Max-Min & The lowest the utility is maximized \\
\cline{2-5}
~ & Cluster balance & Known &  Fair cluster & The ratios of the groups are balanced for each cluster \\
\cline{1-5}

\multirow{2}{*}{Individual} & 
Lipschitz property & 
Known & Fairness through awareness & Similar individuals have similar outcomes \\
\cline{2-5}
~ & Causal mode & Known & Counterfactual fairness & Same prediction for actual/counterfactual individuals \\
\cline{1-5}

Hybrid & 
Bounding & 
Known & Differential fairness & Positive/negative outcomes bounded across sub-groups \\

\bottomrule
\end{tabular}
\end{table}


\begin{itemize}
    \item \textbf{Demographic parity:} The percentages of a positive outcome across different sensitive groups (denoted by $x_s$) should be the same, i.e., $p(\hat{y}=1|x_s=s_i) = p(\hat{y}=1|x_s=s_j)$. In the example of Figure~\ref{fig:twobias}, $40\%$ of females are predicted as positive, while the ratio is $70\%$ for male. Due to the large gap between the two ratios, it is probable that the model violates demographic parity.
    \item \textbf{Equal opportunity:} Different groups should have equal true positive rates, i.e., $p(\hat{y}=1|x_s=s_i,y=1) = p(\hat{y}=1|x_s=s_j,y=1)$. In Figure~\ref{fig:twobias}, three out of six truly positive individuals in the female group are predicted as positive, with a true positive rate of $50\%$; in contrast, the true positive rate of the male group is $86\%$, which is significantly higher than $50\%$. Therefore, it is probable that the model fails to guarantee equal opportunity.
    \item \textbf{Equalized odds:} Different groups should have equal true positive and false positive rates, i.e., $p(\hat{y}=1|x_s=s_i,y=1) = p(\hat{y}=1|x_s=s_j,y=1)$ and $p(\hat{y}=1|x_s=s_i,y=-1) = p(\hat{y}=1|x_s=s_j,y=-1)$. This measurement is more restrictive than demographic parity and equalized odds since we require both true and false positive rates to be the same. It is often used when we strongly care about predicting the positive outcomes correctly.
    \item \textbf{Overall accuracy equality~\cite{berk2021fairness}:} The accuracies across the sensitive groups are the same, i.e., $p(\hat{y}=1|x_s=s_i,y=1) + p(\hat{y}=-1|x_s=s_i,y=-1) = p(\hat{y}=1|x_s=s_j,y=1) + p(\hat{y}=-1|x_s=s_j,y=-1)$. Unlike the above measurements, overall accuracy equality emphasizes both true positive and true negative rates. This can be used in the scenarios where true negatives are as desirable as true positives.
    \item \textbf{Treatment equality~\cite{berk2021fairness}:} The ratios of false negative prediction to false positive prediction should be equal across groups, i.e., $\frac{p(\hat{y}=1|y=-1, x_s=s_i))}{p(\hat{y}=-1|y=1, x_s=s_i))} = \frac{p(\hat{y}=1|y=-1, x_s=s_j))}{p(\hat{y}=-1|y=1, x_s=s_j))}$. This measurement can be used when incorrectly classifying an individual results in a bigger loss.
    \item \textbf{Equalizing disincentives~\cite{jung2020fair}}: The difference between true positive rates and false positive rates should be equal across groups, i.e., $p(\hat{y}=1|y=1, x_s=s_i)) - p(\hat{y}=1|y=-1, x_s=s_i)) = p(\hat{y}=1|y=1, x_s=s_j)) - p(\hat{y}=1|y=-1, x_s=s_j))$. This measurement has strong emphasis on classifying positives both correctly and incorrectly.
    \item \textbf{Rawlsian Max-Min fairness principle~\cite{rawls2001justice}:}  Let $U(\theta, s)$ be the utility of the sensitive group $s$ with weights $\theta$. The Rawlsian Max-Min fairness can be quantified as $\argmax_\theta \min_s U(\theta, s)$. This measurement encourages maximizing the utility of the group with the lowest utility, where the utility metric can be accuracy, AUC, etc. While the definition of this measurement does not explicitly consider sensitive groups, it implicitly treats the individuals with the lowest utility as the minority group. It is commonly used in the scenarios where the sensitive information is unknown~\cite{hashimoto2018fairness,lahoti2020fairness}.
    \blue{\item \textbf{Fair clustering~\cite{chierichetti2017fair}: } It is defined for quantifying unfairness in clustering problems.  Let $N(\hat{y}=c|x_s=s_i)$ denote the number of individuals that are assigned to cluster $c$ with sensitive attribute $s_i$. The balance of the cluster $c$ is defined as the minimum ratios of the numbers of individuals in each group, i.e.,  $\text{balance}(c) = \min \{\frac{N(\hat{y}=c|x_s=s_i)}{N(\hat{y}=c|x_s=s_j)}, \frac{N(\hat{y}=c|x_s=s_j)}{N(\hat{y}=c|x_s=s_i)}\}$. The generated clusters are considered fair if for every cluster $c$, $\text{balance}(c) \ge \delta$, where $\delta$ is a threshold.  }
\end{itemize}

\subsubsection{Individual Fairness Measurements} \hfill\\
Instead of measuring fairness across different sensitive groups, individual fairness measurements consider the fairness for each individual with the intuition that similar individuals should be treated as similarly as possible. Some representative scenarios include:
\begin{itemize}
    \item \textbf{Fairness through awareness~\cite{dwork2012fairness}:} Any two individuals who have similar non-sensitive attributes should receive a similar outcome. Let $d(\textbf{x}_a, \textbf{x}_b)$ define the difference between the attributes of two individuals $\textbf{x}_a$ and $\textbf{x}_b$. If $d(\textbf{x}_a, \textbf{x}_b)$ is small, then $D(\hat{y}_a, \hat{y}_b)$ should also be small, where $D(\cdot, \cdot)$ computes the prediction difference. Formally, the goal is to bound $D(\hat{y}_a, \hat{y}_b)$ and make the model satisfy the $(D, d)-Lipschitz$ property: $D(\hat{y}_a, \hat{y}_b) \le d(\textbf{x}_a,\textbf{x}_b)$. The specific formulation of $D(\cdot, \cdot)$ and $d(\cdot, \cdot)$ are usually determined by the task at hand.

    \item \textbf{Counterfactual fairness~\cite{kusner2017counterfactual}:} It is defined with causal models. Let $(U, V, F)$ be a causal model, where, $U$ denotes the set of background variables, $V$ denotes the set of observable variables, and $F$ denotes the set of functions associated with $V$. $\hat{y}$ is counterfactually fair towards an individual if the prediction is the same for the actual individual and a counterfactual individual that belongs to a different sensitive group, i.e., for any context $\textbf{x}$, $p(\hat{y}_{x_s \leftarrow s_i} (U)=y|\textbf{x}) = p(\hat{y}_{x_s \leftarrow s_j} (U)=y|\textbf{x})$.
\end{itemize}

\subsubsection{Hybrid Measurements} \hfill\\
Some hybrid measurements, instead of measuring the difference in the group- or the individual-level, focus on the subsets of multiple sensitive attributes. We use $\textbf{s}_i$ and $\textbf{s}_j$ to denote two different tuples representing all the sensitive attribute values. For example, if we consider both gender and race in the \emph{Adult} dataset, an instantiation of such tuple can be $\{male, black\}$ or $\{female, white\}$. We introduce a representative measurement as below.
\begin{itemize}
    \item \textbf{Differential fairness~\cite{foulds2020intersectional}:} A model $f_\theta$ is $\epsilon$-differentially fair if $e^{- \epsilon} \le \log \frac{p(\hat{y}=0|\textbf{s}_i)}{p(\hat{y}=0|\textbf{s}_j)} \le e^{\epsilon}$ and $e^{- \epsilon} \le \log \frac{p(\hat{y}=1|\textbf{s}_i)}{p(\hat{y}=1|\textbf{s}_j)} \le e^{\epsilon}$ for all $\textbf{s}_i$ and $\textbf{s}_j$. This criteria states that the predictions are similar and bounded within the range $(e^{-\epsilon}, e^{\epsilon})$ regardless of the combinations of the sensitive attributes.
\end{itemize}


\begin{figure}
    \centering
    \includegraphics[width=\textwidth]{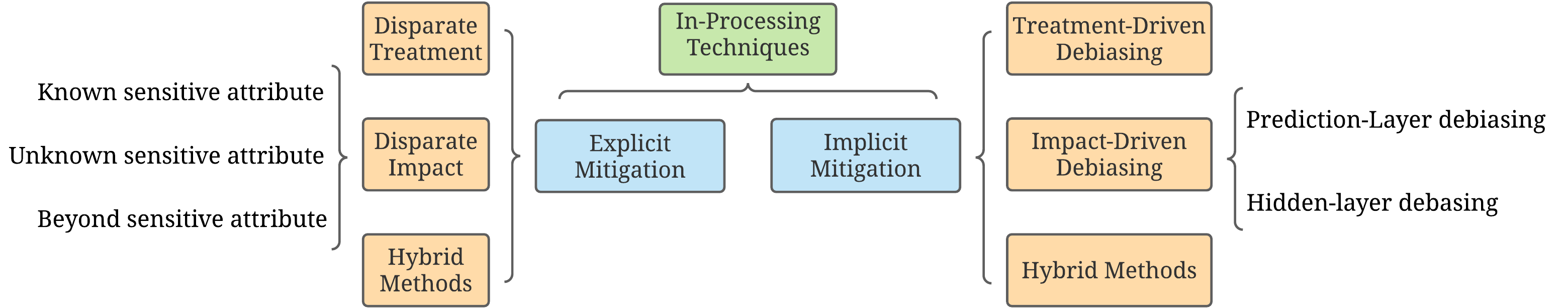}
    \caption{Structure of the survey.}
    \label{fig:taxonomy}
\end{figure}

\subsection{Survey Structure}
\label{sec:2.3}



This survey will give an overview of the recent in-processing techniques developed for various machine learning tasks. We will summarize the core ideas and design principles regardless of the specific applications. We divide existing techniques into two categories:
\begin{itemize}
    \item \textbf{Explicit methods:} Research in this line often focuses on explicitly revising the training objective $L(\mathcal{D};\theta)$. One common strategy is to add fairness related constraints and solve the problem with constrained optimization techniques. Another strategy is adding a regularization term into the objective function to penalize the unfairness behaviors. We provide detailed categorizations in Section~\ref{sec:3}.
    \item \textbf{Implicit methods:} An alternative direction is to debias the latent representation $\textbf{z}$ to implicitly remove bias from the model. The intuition is that if $\textbf{z}$ is \blue{less biased}, the resulting predictions $\hat{y}$ from $\textbf{z}$ will \blue{also be less biased}. We provide detailed categorizations in Section~\ref{sec:4}.
\end{itemize}
We further categorize explicit and implicit methods into several sub-categories based on whether the algorithms are designed to mitigate the disparate impact (i.e., group fairness\blue{/bias}), disparate treatment (i.e., individual fairness\blue{/bias}), or both of them. \blue{Under each sub-category, based on the existing techniques, we either provide a high-level summary of the key ideas (e.g., a unified formula) with some representative instances, or further divide them based on the scenarios that the algorithms can tackle.} An overview of the taxonomy used in this article is summarized in Figure~\ref{fig:taxonomy}.

\blue{\subsection{Strategy for Paper Collection and Category Determination} \label{sec:2.4}}
\blue{We identified the relevant papers and determined the categories based on the following iterative procedure. Firstly, we identified a small set of papers by using the query ``fair machine learning'' along with some related keywords, such as ``bias'', ``discrimination'', and "unfairness", to narrow down the search results. Note that we did not explicitly put the keyword ``in-processing'' in the query since many in-processing papers do not explicitly mention this term in the title or abstract. Instead, we focused on those important papers published on machine learning or data mining conferences or journals, and manually examined them to collect the relevant ones. Secondly, we categorize the papers according to our taxonomy (e.g., explicit and implicit methods).}

\blue{We iteratively repeated the above two steps to collect more papers following the categories determined in the previous iterations, and gradually refined the categories and developed sub-categories based on the newly collected papers. Finally, the identified taxonomy converged to the one shown in Figure~\ref{fig:taxonomy}. Note that rather than desperately trying to cover all the work in this domain, we only selected the most representative papers under each sub-category to make our discussion focused. In essence, some sub-categories (e.g., adversarial debiasing) can have many extensions and variants. In this case, we manually traced the citation graph to identify the most influential papers for discussion.}

\blue{\section{Explicit Unfairness Mitigation}\label{sec:3}}

\begin{figure}
    \includegraphics[width=0.9\textwidth]{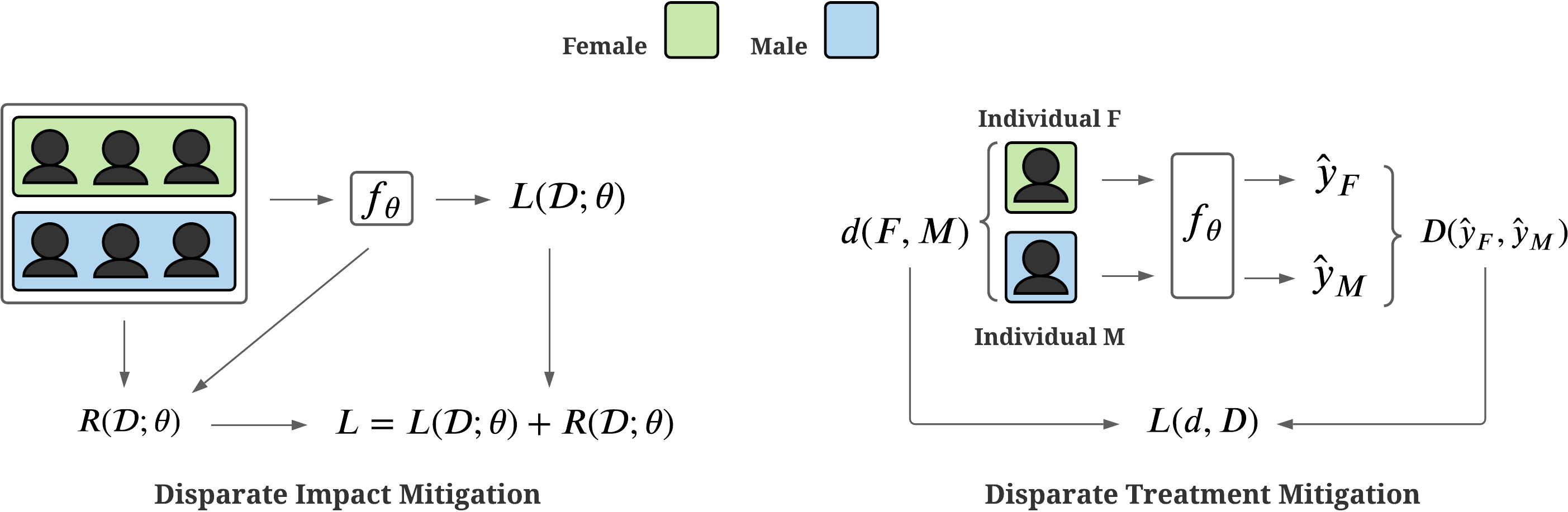}
    \caption{An illustration of explicit mitigation of bias. The left-hand side shows an example of disparate impact mitigation with regularization terms. In addition to the original loss $L(\mathcal{D}; \theta)$, a fairness term $R(\mathcal{D};\theta)$ is formulated to enforce group fairness. The existing studies mainly differ in how $R(\mathcal{D}; \theta)$ is designed. The right-hand side illustrates disparate treatment mitigation. Here, $d(\cdot, \cdot)$ and $D(\cdot, \cdot)$ measure the distance between any two individuals in attribute space and prediction space, respectively. The objective is to minimize the discrepancy between these two distances for individuals with similar non-sensitive attributes but different sensitive attributes. In this example, a female (F) and a male (M) with similar non-sensitive attributes are enforced to have similar predictions.}
    \label{fig:explicitmethods}
\end{figure}

Explicit \blue{unfairness} mitigation can be achieved by explicitly adding fairness regularizers or constraints to the objective functions of machine learning models. The high-level ideas are illustrated in Figure~\ref{fig:explicitmethods}. In this section, we summarize different explicit mitigation methods according to whether they tackle disparate impact or disparate treatment. Additionally, researchers have also studied hybrid methods that address both of them. A summary of the surveyed explicit mitigation methods is provided in Table~\ref{tbl:explicit}.

\subsection{Disparate Impact Mitigation}
\label{sec:3.1}
The existing explicit mitigation methods for addressing disparate impact can be categorized based on whether sensitive attribute information is known. Beyond these, some other studies have also explored fairness problems that are irrelevant to sensitive attributes.


\subsubsection{Fairness with known sensitive attribute information} \label{sec:3.1.1} \hfill\\
A large body of existing work has investigated how to mitigate disparate impact when the sensitive information, such as the race or gender information for each individual, is available in the training data. A typical strategy is to guide the model with fairness regularization so that predictions are less dependent on the sensitive attributes. A general objective function with fairness learning is formulated as:
\begin{equation}
L(\mathcal{D}; \theta)+ \lambda {\left \|\theta\right\|}_{2}^{2} + \eta R(\mathcal{D}; \theta) ,
\end{equation}
where the first term is the loss of the main task (e.g., classification or regression), the second term is a standard regularizer on model parameters, the third term is a designed fairness regularizer, and $\eta$ and $\lambda$ are hyperparameters. Here, the fairness regularizer $\text{R}(\mathcal{D}; \theta)$ is the key to achieve the model fairness.

The core idea of the regularizer design is to quantify and minimize the correlation between the sensitive attributes and the prediction. A representative strategy to quantify such correlation is based on mutual information. \cite{kamishima2011fairness} introduced \emph{prejudice index} (PI in short) to quantify the degree of dependence between a sensitive variable and a target variable. It is defined as the sample distributions over a given sample set of the mutual distribution between the sensitive groups and the predictions. By minimizing the mutual information between these two, we enforce the predictions less dependent on the sensitive attributes. In this sense, the model tends to be fair across groups. Although this paper focuses on simple linear regression and naive Bayesian classifiers, the idea is general and could be also applied to other models. Other notable regularizers include minimizing Wasserstein-1 distances between the classifier outputs and sensitive  information~\cite{jiang2020wasserstein}, absolute correlation regularization that takes a simplified view by minimizing the absolute value of the correlation between the sensitive groups and the predictions~\cite{beutel2019putting}, and residual between the clicked and unclicked item and the members in groups in recommender systems~\cite{beutel2019fairness}, etc.

\begin{table}[t]
\small
\centering
\caption{A summary of explicit mitigation methods. \blue{\emph{Regularization} refers to adding a penalty term to the objective function to make the optimal solution fairer. \emph{Constraint optimization} refers to optimizing the parameters with the original objective in the presence of some (hard) constraints on the parameters. Note that regularization is also regarded as a soft constraint in the literature (i.e., a constraint which is preferred but not required to be satisfied). In this survey, we distinguish these two by grouping the papers with hard constraint into constraint optimization and papers with soft constraint into regularization.}  }
\label{tbl:explicit}
\begin{tabular}{l|l|l|l} \toprule
\textbf{Fairness Goal} & \textbf{Training Scheme} & \textbf{Sensitive Attribute} & \textbf{Method} \\
\midrule

\multirow{7}{*}{Disparate impact} & 
\multirow{3}{*}{Regularization}  & 
\multirow{3}{*}{Known} & Prejudice index~\cite{kamishima2011fairness}, Absolute correlation ~\cite{beutel2019putting}\\
~ & ~ & ~ &   Wasserstein fair ~\cite{jiang2020wasserstein}, Pairwise comparisons~\cite{beutel2019fairness}\\
~ & ~ & ~ & Fair regression~\cite{agarwal2019fair}, Fair decision trees~\cite{aghaei2019learning}\\
 \cline{2-4}

~ & \multirow{4}{*}{Constraint optimization} & 
\multirow{2}{*}{Known} & Flexible mechanism~\cite{zafar2017fairness}, Tractable constraints~\cite{zafar2019fairness},\\
~ & ~ & ~ & Convex-concave~\cite{shen2016disciplined}\\

\cline{3-4}
~ & ~ &  Unknown &  DRO~\cite{hashimoto2018fairness}, ARL~\cite{lahoti2020fairness}, Proxy Fairness ~\cite{gupta2018proxy}\\
\cline{3-4}
~ & ~ & Beyond &  Paper matching ~\cite{kobren2019paper}\\ \cline{1-4}

\multirow{4}{*}{Disparate treatment} & 
\multirow{2}{*}{Regularization} & 
\multirow{2}{*}{Known} & Controlled direct effect~\cite{di2020counterfactual}, Fair decision trees~\cite{aghaei2019learning}, \\
~ & ~  & ~ & Convex fair regression~\cite{berk2017convex}\\
\cline{3-4}
 \cline{2-4}

~ & \multirow{2}{*}{Constraint optimization} & 
\multirow{2}{*}{Known} & Fairness Through Awareness ~\cite{dwork2012fairness}, Logit pairing ~\cite{garg2019counterfactual}\\
~ & ~  & ~ & Counterfactual fairness~\cite{kusner2017counterfactual}\\
\cline{3-4}

\cline{1-4}
\multirow{2}{*}{Hybrid methods}& \multirow{2}{*}{Constraint optimization} & \multirow{2}{*}{Known}&  subgroup fairness~\cite{kearns2018preventing}, rich subgroup fairness\cite{kearns2018preventing}\\
~ & ~ & ~ &  Maxmin-Fair Ranking~\cite{garcia2021maxmin}\\

\bottomrule
\end{tabular}

\label{tab:explicit}
\end{table}

Another idea of designing objective functions is to minimize the loss under fairness constraints, which can be formulated as a constraint optimization problem:
\begin{equation}
\begin{aligned}
\min_{\theta} \quad & L(\mathcal{D}; \theta)+ \lambda {\left \|\theta\right\|}_{2}^{2}\\
\textrm{s.t.} \quad & \Omega(\mathcal{D}; \theta) < 0, \\
\end{aligned}
\end{equation}
where $L(\mathcal{D}; \theta)$ is the loss of the main task, the second term is a standard regularizer, and $\Omega$ is a fairness constraint that is usually defined as a convex function. An example is to formulate $\Omega$ as the covariance between the sensitive attributes and the signed distance from the feature vectors to the decision boundary~\cite{zafar2017fairness,zafar2019fairness}. In this example, $\Omega$ is a convex function because the signed distance is convex with respect to $\theta$. The convex nature ensures that $\Omega$ will not increase the complexity of the training.

Complementing the above strategies that mainly focus on model predictions, some other studies alternatively investigate the scenarios where ground truth is available in the historical data. A noteworthy example is mitigating \emph{disparate mistreatment}~\cite{zafar2017fairness2}, which aims to achieve similar misclassification rates for different sensitive groups. Disparate mistreatment is usefull in many real-world scenarios, such as assigning risk scores to criminal offenders. For example, ProPublica\footnote{\url{https://github.com/propublica/compas-analysis}} collects data of criminal offenders in Broward County, Florida, during
2013-2014. The data consists of offenders' demographic features, such as gender, race, and age, as well as the offenders' criminal history. COMPAS (Correctional Offender Management Profiling for Alternative Sanctions) tool\footnote{\url{https://www.documentcloud.org/documents/2702103-Sample-Risk-Assessment-COMPAS-CORE.html}} is then used to assign a risk score to each offender. The data also provides ground truth on whether the offenders have recidivated within two years. An individual is misclassified if a high (low) risk score is assigned to an offender who has not (has) recidivated within two years. The automated decision making system may suffer from disparate mistreatment if the misclassification rates for different sensitive groups (e.g., race) differ. To address this, a constraint optimization problem is formulated to bound the difference of the misclassification rates between two sensitive groups. The problem can then be efficiently solved with convex-concave programming~\cite{shen2016disciplined}. In this way, the classifier learns the optimal decision boundary \blue{subject to the fairness constraints}, which \blue{facilitates} fair treatments across sensitive groups. \blue{However, such constraints only bound the overall misclassification rates, while specific individuals within a group can still be treated unfairly.}

\blue{Recent studies have explored fairness in unsupervised learning tasks, where the label information is unavailable. The existing studies in this line of research mainly focus on clustering, which is often considered as the most common unsupervised learning problem. One representative work, Fair Clustering~\cite{chierichetti2017fair}, defines fairness in clustering under the disparate impact doctrine. Specifically, it defines the balance of a cluster as the ratio between the numbers of individuals belonging to each sensitive group within the cluster. A clustering algorithm is considered fair if all the generated clusters are relatively balanced. Unfortunately, achieving the optimal solution under such fairness definition is NP-hard because the number of combinations grows exponentially with more individuals. This work develops an approximation algorithm based on \textit{fairlets}, which refers to small and fair clusters. The key idea is to first obtain some farilets, which can be solved in polynomial time by relating it to the graph covering problem. Then it clusters the fairlets using standard clustering algorithms. Since the fairlets are balanced, the obtained clusters (each cluster is formed by multiple farilets) will also be balanced. A follow-up work extends Fair Clustering by making the first step (i.e., fairlets decomposition) more efficient by mapping the data points into a tree metric. We refer interested readers to~\cite{backurs2019scalable} for more details.   }


\subsubsection{Fairness without sensitive attribute information} \label{sec:3.1.2} \hfill\\
All the above studies assume that sensitive attributes, such as gender and race, are available in the dataset. However, collecting sensitive attributes itself is sometimes infeasible due to privacy issues. To deal with these situations, researchers have investigated how to mitigate \blue{unfairness} when we do not have full access to sensitive attributes. The main \blue{idea} of these studies is that, while we do not have direct access to the unobserved sensitive attributes, they are often correlated with some other observed attributes. For example, the races could be correlated with the zip codes~\cite{datta2017proxy}. As such, we could mitigate the racial \blue{unfairness} even without knowing the races by treating the individuals with a certain zip code as the minority group. \blue{This is because if zip codes can divide the individuals in a similar way as races, mitigating unfairness with zip codes will have similar effects as races. Note that whether this idea is effective highly depends on the degree of the correlations between the sensitive attribute and the other attributes. Some attributes could have strong correlations with the sensitive attribute while the others may only have weak correlations. For example, it is well known that vocal pitch correlates with gender. In this case, the vocal pitch could serve as an almost perfect substitute for gender. In contrast, in the case of races and zip codes, while they have a good correlation in general, the exact correlation may differ in different cities and countries. Thus, a specific zip code may not necessarily correspond to exactly one race. The degree of correlations will determine the fairness performance of the mitigation algorithms since they highly rely on such correlations to achieve fairness.}

\emph{Proxy Fairness}~\cite{gupta2018proxy} explores this idea by addressing fairness on proxy groups, i.e., the groups identified based on the correlated observed attributed, as substitutes for the sensitive groups. In the above example, we can define the proxy groups based on the zip code as substitutes for races. In this way, any existing fairness mitigation techniques could be applied to the proxy groups, which could also indirectly address the fairness issues in the true sensitive groups. In~\cite{gupta2018proxy}, they formulate a constraint optimization problem on the proxy groups and surprisingly observe that such a simple strategy could work well in practice. However, the authors also note that the effectiveness of the idea depends on the choice of fairness metric and the alignment between the proxy groups and the true sensitive groups. This approach requires careful selection of the proxy features and could only meet some specific needs.

Another direction focuses on improving Rawlsian Max-Min Fairness~\cite{rawls2001justice}, which aims at maximizing the utility of the worst-case group, i.e., the group with the lowest utility. The methods in this line of research also rely on the correlated observed attributes for \blue{unfairness mitigation}. The main idea is to identify the regions with higher classification errors using the observed attributes and give a higher weight to the individuals within these regions to optimize the worst-case performance. One noteworthy training strategy is based on \emph{Distributionally Robust Optimization} (DRO)~\cite{hashimoto2018fairness}. The assumption behind is that the performance of minority groups is sacrificed when achieving the overall performance since the contribution of minority groups to the overall loss is significantly less than that of majority groups, that is, the minority group is underrepresented during the training. To address this issue, DRO minimizes
\begin{equation}
    \mathbb{E}_P [L(\mathcal{D};\theta) - \eta]_+^2,
\end{equation}
where $L(\mathcal{D};\theta)$ is the classification loss, $[\cdot]_+$ drops the instances with loss smaller than $\eta$, and $\eta$ is a dual variable controlling the threshold for small loss. This objective function can optimize the worst-case performance in that both the dual variable $\eta$ and the squared term can up-weight the samples with high losses. However, DRO may suffer from the risk of focusing on optimizing the performance of outliers since the losses of outliers are usually above the threshold.

A follow-up work proposes to address this limitation with adversarial reweighted learning~\cite{lahoti2020fairness}. In this work, the Rawlsian Max-Min Fairness objective is formulated as a minimax problem of a zero-sum game between two players, where one player tries to minimize the reweighted loss function, and the adversary player aims to maximize the loss by assigning the weights. The weight assignment is achieved by a neural network that takes as input the observed features and outputs a value between 0 to 1 as the weight. During the training process, the adversary tends to assign higher weights to the instances with higher losses, and the other player will focus more on these instances to optimize the worst-case performance. A linear adversary is used and shown to deliver the best performance since it can mitigate the impact of the noisy outliers.

\subsubsection{Beyond sensitive attributes} \label{sec:3.1.3}\hfill\\
While previous research efforts mainly center on sensitive attributes, there exist fairness scenarios that are irrelevant to sensitive attributes and instead focus on the fairness issues in specific application domains. \blue{We note that such fairness problems can be very specific may not be encountered in other domains.}

Paper matching problem~\cite{kobren2019paper} is a typical scenario that falls into this subcategory. It aims at automatically matching reviewers to the papers in the reviewing process. Instead of addressing fairness problems across sensitive groups, they focus on fairness regarding papers and reviewers. Specifically, they identify two fairness goals in paper reviewing: (1) the reviewers assigned to a paper should collectively possess sufficient expertise, and (2) each reviewer should have a reasonable workload in terms of the number of the assigned papers. On one hand, each paper is treated as an individual, and the assignment should be in fair in terms of the expertise of the assigned reviewers. On the other hand, each reviewer is expected to be fairly treated in terms of workloads. Both of these fairness goals do not involve sensitive information . The paper matching problem can be formulated as a global optimization problem that maximizes the sum of affinities, which are scores indicating the affinity of reviewer-paper pairs. Then, the two fairness goals are achieved via adding constraints to the optimization process.

\subsection{Disparate Treatment Mitigation}
\label{sec:3.2}
Disparate treatment mitigation can be achieved by penalizing the discrimination towards individuals with different sensitive attributes in objectives, i.e., similar individuals in different groups should be treated as similarly as possible. The methods in this research line often rely on task-specific metrics to measure the similarity between individuals. The idea is applicable to both classical machine learning models and deep models, where fairness is usually achieved by formulating a constrained optimization problem or adding regularization terms to the loss functions.

\emph{Fairness Through Awareness}~\cite{dwork2012fairness} is a representative approach for traditional classifiers. The key idea is to formulate a fair classifier as a constrained optimization problem, which minimizes the (linear) classification loss subject to a fairness constraint. Specifically, given a distance metric between individuals $d(\cdot, \cdot)$ and a distance measure between outputs $D(\cdot, \cdot)$, for any pair of individuals $\textbf{x}_a$ and $\textbf{x}_b$, the model is optimized subject to $(D, d)-Lipschitz$ property:
\begin{equation}
   D(f_{\theta}(\textbf{x}_a), f_{\theta}(\textbf{x}_b)) \le d(\textbf{x}_a,\textbf{x}_b),
\end{equation}
where $f_\theta$ is the model and $f_\theta(\textbf{x})$ denotes the prediction result of $\textbf{x}$. The above constraint expresses that the distance of the outputs is bounded by the distance of the corresponding inputs. \blue{In this way, similar individuals (i.e., the distance of inputs is small) will also have similar outcomes (i.e., the distance of outputs is also small since it is bounded by the distance of inputs), which will lead to individual fairness.} If $f_\theta$ is linear, the optimization problem can then be expressed as a linear programming that can be solved efficiently. Here, the two distance metrics $D(\cdot, \cdot)$ and $d(\cdot, \cdot)$ are key to achieve the goal of ``treating similar individuals similarly". Since $f_\theta(\textbf{x})$ is essentially a probability distribution, a straightforward choice for $D(\cdot, \cdot)$ is a statistical distance metric between two probability distributions $P$ and $Q$ on a finite domain $A$ (i.e., there are $A$ possible outcomes), denoted by
\begin{equation}
    D_{\text{tv}}(P,Q) = \frac{1}{2} \sum_{a \in A} |P(a)-Q(a)|,
    \label{eq:5}
\end{equation}
where \blue{$D_{\text{tv}}(P,Q)$ is the total variation distance, and} two very different distributions tend to have larger statistical distances and vice versa. $d(\cdot, \cdot)$ is a task-specific metric describing the similarity level of two individuals, which highly depends on the task.

An alternative idea for disparate treatment mitigation is \emph{counterfactual fairness}~\cite{kusner2017counterfactual}, which leverages the causal framework to model the sensitive attributes. A decision is considered fair to an individual if the prediction is the same for the actual and counterfactual worlds. One representative example of counterfactual fairness modeling is counterfactual logit pairing~\cite{garg2019counterfactual}, which is a robustness term that penalizes the norm of the logit difference between a pair of an actual example and their counterfactual example. Another similar work proposes to remove the direct effect of the sensitive attribute with regularization~\cite{di2020counterfactual}, which aims to minimize the difference between an actual individual and a counterfactual individual that belongs to a different sensitive group.


Other studies have designed constraints or regularizers for different kinds of tasks or models with linear or non-linear mappings. Some representative examples include adding fairness regularizers for achieving fairness in regression tasks~\cite{berk2017convex}, and learning fair decision trees~\cite{aghaei2019learning}.

\subsection{Hybrid Methods for Explicit Bias Mitigation}
\label{sec:3.3}
Disparate impact mainly addresses group-level fairness issues but could not handle individual-level ones. Similarly, disparate treatment focuses on individual-level rather than group-level fairness. However, there are hybrid scenarios that fall beyond these two categories. This subsection reviews two representative ones: subgroup fairness that defines fairness in the subgroup-level and hybrid methods that attempt to achieve both group- and individual-level fairness.

Rather than defining fairness purely at the group- or individual-level, subgroup fairness~\cite{kearns2018preventing} defines fairness over a combinatorially large number of subgroups. \cite{kearns2018preventing} gives a toy example as a failure of group-level fairness, called fairness gerrymandering. Consider a problem with two sensitive attributes including race and gender\blue{, where men/women and whites/blacks are equally represented in the target population.} Suppose a binary classifier predicts positive if and only if an individual is a black man or a white woman, then the positive rate is $50\%$ within each sensitive group. That is, the classifier will be considered to be fair under a disparate impact metric; however, it is unfair to black men or white women. Motivated by this example, \cite{kearns2018preventing} proposes to address subgroup fairness, where the classifier is expected to satisfy the fairness constraints for each possible subgroup. The proposed solution formulates the problem as a two-player zero-sum game between a Learner and an Auditor, where the Learner's decision space is the classifier itself, and the Auditor has the decision space of subgroups. This game can be solved with Fictitious Play with provably asymptotic convergence. Their follow-up work~\cite{kearns2019empirical} further empirically shows that this approach works well in various datasets and demonstrates that subgroup fairness can be satisfied with the reasonable cost in terms of computational resources and accuracy. Unlike group-level or individual-level fairness, this research line focuses on fairness at the subgroup-level. Intuitively, subgroup fairness will reduce to group-level fairness if there is only one subgroup and will become individual-level fairness if there is only one individual for each subgroup.

One idea to achieve both disparate impact and disparate treatment is based on constrained optimization. A notable example is \emph{Maxmin-Fair Ranking}~\cite{garcia2021maxmin}, which aims to optimize individual fairness under group fairness constraints. Specifically, this paper focuses on a ranking problem, where people or items are assigned scores that indicate the relevance to the query. It is important that the ranking system should be fair since the result will have a direct tangible impact on the people or items being ranked. This paper formulates a minimax problem to minimize individual unfairness while enforcing the group fairness constraints that are derived from sensitive attributes:
\begin{equation}
    r' \in \argmax_{r \in \mathcal{S}} \min_{\textbf{x} \in \mathcal{X}} V(r, \textbf{x}),
\end{equation}
where $r$ or $r'$ denotes a ranking of the individuals, $\mathcal{S}$ denotes the subset of the rankings that satisfy the group fairness constraints, $V(r, \textbf{x})$ denotes the utility of $\textbf{x}$ if we place $\textbf{x}$ according to rank $r$; that is, our goal is to maximize the worst-case utility across individuals under the group fairness constraints. Based on this formulation, the paper further improves the individual treatment with randomization, i.e., deriving a probability distribution over valid rankings instead of obtaining a deterministic solution. The optimization can be solved with combinatorial optimization in polynomial time.

\begin{figure}
    \includegraphics[width=\textwidth]{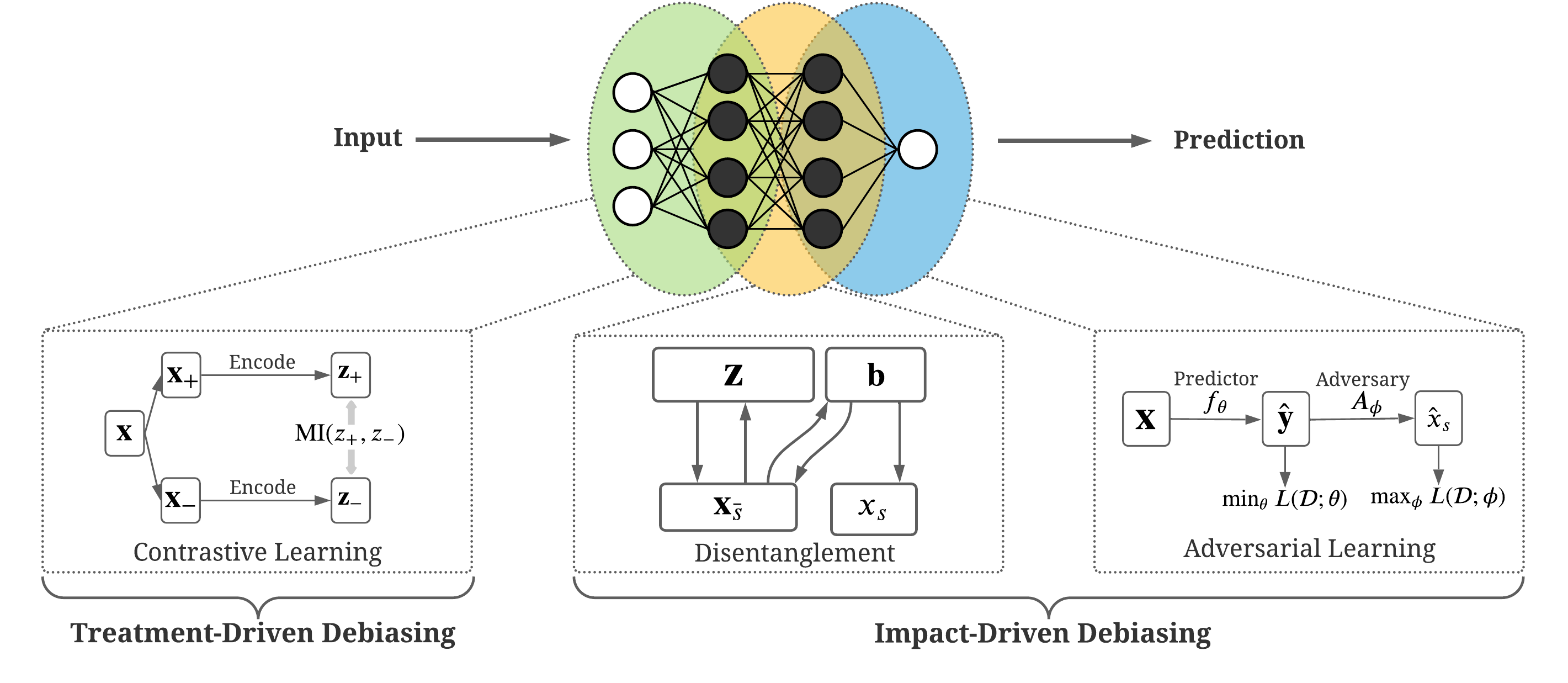}
    \caption{Some representative techniques for implicit mitigation of bias, where different techniques often focus on different phases of the model. Contrastive learning methods mitigate disparate treatment by minimizing the mutual information between the representations obtained from positive samples ($\textbf{x}_+$) and negative samples ($\textbf{x}_-$), which have similar non-sensitive attributes but different sensitive attributes. Disentangled learning approaches mitigate the disparate impact by learning disentangled sensitive and non-sensitive representations and using the obtained non-sensitive representations for downstream applications. Adversarial learning strategies use an auxiliary adversary network to minimize the predictive probability of the sensitive attributes from the representations and mitigate the bias extracted in the representation.}
    \label{fig:implicit}
\end{figure}

\blue{\section{Implicit Unfairness Mitigation}\label{sec:4}}

Implicit \blue{unfairness} mitigation refers to the algorithms which detect and eliminate discrimination via representations. These approaches are mainly designed for deep models, where learning representations is crucial. \blue{They seek to remove bias from the learned representations such that the resulting predictions will also be less biased. Note that, as discussed in Section~\ref{sec:2}, an unbiased prediction may not necessarily be fair. Nevertheless, in most cases, it has been reported in many implicit mitigation methods that removing bias aligns with the common fairness goals~\cite{zhang2018mitigating,creager2019flexibly,cheng2021fairfil}. } Figure~\ref{fig:implicit} illustrates some representative techniques for implicit mitigations, which can be grouped into impact-driven \blue{methods} and treatment-driven \blue{methods}. Some representative techniques for impact-driven \blue{methods} include adversarial learning and disentangled representation learning, where the former uses an adversary to remove the sensitive information from the representations, and the latter decorrelates the sensitive and non-sensitive information to preserve \blue{unbiased} representations for downstream applications. Alternatively, contrastive learning achieves treatment-driven debiasing by enforcing similar representations for positive and negative sample pairs. In addition, some hybrid methods have been proposed to tackle both disparate impact and disparate treatment. A summary of the implicit mitigation methods is tabulated in Table~\ref{tab:implict}.


\begin{table}[t]
\centering
\caption{A summary of implicit mitigation methods.}
\begin{tabular}{l|l|l} \toprule
\textbf{Goal} & \textbf{Training Techniques} & \textbf{Method} \\
\midrule

\multirow{7}{*}{Impact-Driven} & 
Adversarial Learning & Adversarial Debiasing~\cite{zhang2018mitigating}\\

~ & (Prediction-layer) & Adversarial Recidivism Application ~\cite{wadsworth2018achieving}\\
\cline{2-3}
~ & \multirow{2}{*}{Adversarial Learning} & Censored Representation ~\cite{edwards2015censoring}\\ 
~ & \multirow{2}{*}{(Hidden-layer)} & Transferable Representations ~\cite{madras2018learning}\\
~ & ~ &  Re-embeds word vector~\cite{sweeney2020reducing}, Fair Word Embedding ~\cite{elazar2018adversarial}\\
\cline{2-3}
~ & 
Disentanglement & Flexibly Fair~\cite{creager2019flexibly}\\
~ & (Hidden-layer)&  Disentangled Representations~\cite{locatello2019fairness}\\
\cline{1-3}

\multirow{3}{*}{Treatment-Driven} & 
\multirow{3}{*}{Contrastive Learning}&
Contrastive Debiasing~\cite{cheng2021fairfil}, Multi-CLRec ~\cite{zhou2021contrastive}\\ 
~ & ~ & Conditional Contrastive ~\cite{tsai2021conditional}, Fair Graph ~\cite{kose2021fairness}\\
~ &  ~ & Fairness-aware Data Augmentation~\cite{kose2021fairness}\\
\cline{1-3}

\multirow{2}{*}{Hybrid methods} & Disentanglement & Flexibly Fair~\cite{creager2019flexibly}, Intersected Fair~\cite{park2021learning} 
\\
\cline{2-3}
~ & Adversarial Learning & Fair Graph Embeddings ~\cite{bose2019compositional}\\

\bottomrule
\end{tabular}

\label{tab:implict}
\end{table}

\subsection{Impact-Driven Debiasing}
\label{sec:4.1}

The goal of impact-driven debiasing is to obtain representations that cannot infer sensitive attributes. As such, the resultant predictions will \blue{be less} biased towards any sensitive groups. Based on which parts of the representations are targeted, the existing studies mainly fall into prediction-layer debiasing and hidden-layer debiasing.

\subsubsection{Prediction-Layer debiasing} \label{sec:4.1.1} \hfill\\
Prediction-layer debiasing aims at mitigating the bias at the prediction layer. This idea is often achieved in a ``backward" way through adversarial learning. The optimization problem is formulated as a minimax game between a predictor and an adversary, where the goal is to maximize the predictor's ability to predict the labels based on the representations and minimize the adversary's ability to predict the sensitive attributes from the representations. In this sense, accurate predictions are achieved without relying on sensitive attribute information.

Consider a predictor $f_\theta$ that is trained to accomplish the task of predicting $\hat{y}$ given $\textbf{x}$ with the loss function $L(\mathcal{D};\theta)$ using a gradient-based method (illustrated in the right part of Figure~\ref{fig:implicit}). The output layer of the predictor then serves as the input of another adversary network $A_\phi$ that aims to predict sensitive attributes $x_s$. The adversary is jointly trained with the predictor to ensure that the predictor satisfies the desired fairness condition but still performs well on the prediction task. More formally, the objective can be written as the following minimax problem:
\begin{equation}
    \min_{\theta} \max_{\phi} L(\mathcal{D};\theta) + L(\mathcal{D};\phi), \label{eqn:adversarial}
\end{equation}
where $L(\mathcal{D};\phi)$ is the adversarial loss that indicates the prediction error of sensitive attributes. 

A representative work~\cite{zhang2018mitigating} has investigated this idea and presented a general formulation of using adversarial learning to mitigate the unwanted bias. This paper shows that the framework can be used to achieve various fairness conditions by adjusting the inputs of the adversarial network based on the fairness definitions. To achieve demographic parity, the adversary will predict the sensitive attributes from the prediction $\hat{y}$ since demographic parity only cares about the percentages of a positive outcome. The adversary can get both $\hat{y}$ and the ground truth $y$ for equalized odds because it relies on the ground truth to calculate true positive and false positive rates. For equal opportunity, we can only feed the training data with $y=1$ to the adversary so that only the true positive rates are considered. This work demonstrated the effectiveness of such adversarial learning framework in mitigating the racial \blue{unfairness} on income prediction tasks. \blue{Note that, although the adversarial network designs are tailored for the fairness measurements, we still use the term \emph{debaising} since the core principle is still enforcing the representations to be less biased.}

A follow-up work further verified the generality of the adversarial learning framework by applying it to criminal history datasets~\cite{wadsworth2018achieving}. The task is to predict the offenders' recidivism based on the attributes of the offenders, which is widely adopted for a decision-making system in the USA Criminal Justice. It is found that there are severe racial biases in criminal history datasets. They use an adversarial network to predict race from the recidivism. The adversarial training penalizes the prediction network if the race is predictable from the recidivism prediction. By adding this adversary network, we can counteract racial \blue{discrimination} in the criminal history dataset. \blue{It is reported that, compared with the standard recidivism prediction model, the adversarial training can decrease the false positive gap from 0.05 to 0.01, and false negative gap from 0.27 to 0.02, with no clear drop of accuracy~\cite{wadsworth2018achieving}.}

\subsubsection{Hidden-Layer debiasing with adversarial learning} \label{sec:4.1.2} \hfill\\
Hidden-layer debiasing aims to mitigate the bias on latent representations learned in intermediate layers. A representative work is learning \emph{censored representation }~\cite{edwards2015censoring}. The model consists of four components: an encoder, a decoder, an adversary, and a predictor. The encoder maps input $\textbf{x}$ to a latent representation $\textbf{z}$ while the decoder takes as input $\textbf{z}$ and the sensitive attributes to reconstruct $\textbf{x}$. The adversary tries to predict the sensitive attributes from latent representation $\textbf{z}$. Finally, the predictor predicts the output based on $\textbf{z}$. The objective can be formulated as 
\begin{equation}
    \min_{\theta} \max_{\phi} L(\mathcal{D};\theta) + L(\mathcal{D};\phi) + L_{\text{r}} (\mathcal{D};\theta), \label{eqn:adversarial2}
\end{equation}
where $L(\mathcal{D};\phi)$ denotes the adversarial loss, and $L_{\text{r}} (\mathcal{D};\theta)$ denotes the reconstruction loss. Without $L_{\text{r}} (\mathcal{D};\theta)$, the objective is very similar to the adversarial learning used in prediction-layer debiasing in Eq.~\ref{eqn:adversarial} with the only difference that the input of the adversary is a hidden-layer. The parameters are optimized with alternate gradient descent and gradient ascent steps. In the first step, the adversary is fixed, and we update the weights of the encoder, the decoder, and the predictor with gradient descent. Then, the adversary takes a step to maximize the loss with the other components fixed. In this way, the representation converges to the point that removes sensitive information.

A follow-up work explores learning adversarially fair and transferable representations~\cite{madras2018learning}. They assume a model with similar components, including an encoder, a decoder, an adversary, and a predictor network. Unlike~\cite{edwards2015censoring}, they address the specific choices of adversarial objectives so that the representations can more closely align with the \blue{debiasing} goals. In addition to classification tasks, they also demonstrate that the method can remove the bias in transfer learning. It has been shown that this method has theoretical grounding for certain \blue{debiasing} objectives and is effective in classification and transfer learning tasks.

\blue{The idea of adversarial debiasing has been extended and explored on unsupervised tasks. A notable example is Deep Fair Clustering~(DFC)~\cite{li2020deep}. The key idea is to separately conduct the clustering algorithm on each sensitive group to obtain pseudo assignments of the instances. These pseudo assignments can be treated as the predicted ``classes''. In this way, we can apply adversarial debiasing designed for supervised learning to the clustering problem. Their empirical results suggest this simple adaptation of adversarial debiasing to clustering problems can make the generated clusters significantly more balanced.}

Recently, hidden-layer debiasing methods have also been developed for sentiment analysis in natural language processing. Demographic identity terms, i.e., the words that can reflect the demographical information of an individual, can show unfair sentiment polarity. For example, the national origin terms like American, Mexican, or
names that tend to belong to African American demographics like
Darnell should be neutral with respect to sentiment. However, these words are shown to have positive or negative sentiment in the embedding space.  To address this problem, in ~\cite{sweeney2020reducing}, the authors remove the discriminated sentiment correlations from the word embedding through adversarial training, which re-embeds word vectors without distorting the meaning. Intuitively, this regime minimizes the ability of the adversary to predict the sentiment polarity while maximizing the learner's ability to preserve the word vector after debiasing. By combining these two objectives in gradient update, they obtain the debiased word embeddings.
Another work has explored similar methods for fair word embedding~\cite{elazar2018adversarial}. This work focuses on reaching text-driven representations that are blind to the attributes we wish to protect. They show that adversarial training is effective for debiasing but may introduce instability in training, which makes the adversary score untrusted during the whole process. Potential directions to improve the results include tuning the capacity and weights of the adversary and using several adversaries as a combination.

\subsubsection{Hidden-Layer debiasing with disentanglement}\label{sec:413} \hfill\\ 
Disentangled representation learning is another strategy for hidden-layer debiasing. The main assumption is that the sensitive and non-sensitive attributes are entangled in the generated representation by an underlying mixing mechanism. To make the prediction results fair across sensitive groups, disentangled representation learning aims at extracting the sensitive information out of the representation and preserving the effective, informative and \blue{unbiased} representation for downstream applications. Disentangled representation learning often considers a generative model that takes the form~\cite{pearl2009causality}:
\begin{equation}
    p(\textbf{x}, \textbf{z}) = p(\textbf{x}|\textbf{z}) \prod_i p(z_i),
\end{equation}
where it is assumed that the observation $\textbf{x}$ is controlled by $l$ independent factors $z_1, z_2, ..., z_l$. The goal of disentangled representation learning is to learn the $\textbf{z}$ such that each dimension in $\textbf{z}$ corresponds to no more than one semantic factor that controls the variation of the data. For example, in image analysis, one dimension of $\textbf{z}$ could extract the eyeglasses of a human face, and the other dimensions represent other parts of the human face~\cite{shen2020interfacegan}. Generally, disentangled representation learning has shown advantages in various applications in terms of its interpretability and generalizability of models~\cite{bengio2013representation}. 

In the context of model fairness, the existing work has studied disentangled representation learning in two settings with or without sensitive attributes. On one hand, disentangled representation learning is an effective way to achieve fairness with known sensitive attributes~\cite{creager2019flexibly}. Consider the encoder and decoder structure in the middle part of Figure~\ref{fig:implicit}, and let $\textbf{b}$ denote the sensitive latent factors. The model training objective is to learn the encoder and decoder while encouraging low \text{MI}(\textbf{b}, \textbf{z}), where $\text{MI}(\cdot, \cdot)$ denotes mutual information. The model is rewarded for decorrelating the latent dimensions of $\textbf{b}$ and $\textbf{z}$ from each other. The model can be trained in an end-to-end fashion by combining a prediction loss. One benefit of this approach is that it \blue{is very flexible}; we can consider multiple sensitive attributes in training and \blue{debias} with different subsets of the sensitive attributes at the test time.

On the other hand, the effectiveness of disentangled representation learning has also been demonstrated even when the sensitive attributes are unobserved in the training data~\cite{locatello2019fairness}. \blue{It applies standard disentangled representation techniques to the model without considering sensitive attributes and conducts a large number of experiments to study the relationship between the disentanglement scores and fairness. Specifically, the standard disentangled representation learning will encourage each factor (i.e., an element in the representation vector) of the learned representation to be semantically independent of the other factors. Here, the specific semantic depends on the context of the task. For example, for the task of learning disentangled face representations~\cite{shen2020interfacegan}, disentanglement aims to enable each element of the representation vector to control a different facial attribute, such as the skin color, hair length, sunglasses color, etc. The disentangled representations have many desirable properties, such as interpretability (i.e., we can explain each facial attribute based on the disentangled representations)~\cite{adel2018discovering}, and better generalization ability (the disentangled representations could capture domain invariant features and be less susceptible to overfitting)~\cite{cai2019learning}. The experiments of \cite{locatello2019fairness} show that the disentangled representations obtained by the standard disentangled representations learning procedure can improve the fairness in the downstream task. They surprisingly found that disentanglement is consistently correlated with increased fairness. A potential reason for why it works is that disentanglement makes the non-sensitive semantic factors less dependent of the sensitive attributes, which inherently makes the predictions less dependent on the sensitive attributes.} Since the sensitive attributes are unobserved in training, the method may help us to avoid biases that we are not aware of.


\subsection{Treatment-Driven Debiasing}
\label{sec:4.2}
The aim of treatment-driven debiasing is to generate similar representations for inputs with similar non-sensitive attributes but different sensitive attributes. Unlike impact-driven debiasing that mainly decorrelate sensitive information from models for group-level fairness, treatment-driven debiasing focuses on fairness modeling at the individual-level. The existing treatment-driven debiasing methods mainly target fairness issues from individual comparison. for example, ``Is it fair that A is chosen but not B?". This is often achieved in a ``forward" way with a contrastive learning framework, where the model is trained to generate similar and dissimilar representations for the positive and negative sample pairs, respectively, to reduce the effect of sensitive attributes. The positive sample pairs are often the ones with the same sensitive attributes but different non-sensitive attributes, while the negative sample pairs often have the same or similar non-sensitive attributes but different sensitive attributes. The positive and negative samples can be either selected from the training data or constructed with data augmentation.


A representative work leverages data augmentation and contrastive learning to debias pre-trained text encoders~\cite{cheng2021fairfil}. Specifically, for each sentence in the training data, another sentence with the same semantic meaning but in a different bias direction is generated. Then a contrastive learning loss is used to maximize the mutual information between the representations of two sentences. Additionally, a regularizer is used to minimize the mutual information between the obtained embedding and the sensitive words. Another example uses conditional contrastive learning~\cite{tsai2021conditional} to remove the effect of sensitive attributes in self-supervised representations. Specifically, the goal of conditional contrastive learning is to obtain similar and dissimilar representations for conditionally-correlated and conditionally-unrelated data, respectively. Here, the condition refers to sensitive attributes. We consider an example of removing gender information from the learned representations. Assuming that we condition on female, the positive pair are data from a female and the corresponding representation from the same female, and the negative pairs are the data from a female and the representations of another female. The positive and negative samples for males can be chosen in a similar fashion. The representations trained in this way will only learn to distinguish the inputs from the same gender but not across genders, which could exclude the gender information from the learned representations. The debiased representations can then be used in other downstream tasks without the undesirable sensitive attributes information.

Some studies have designed contrastive learning strategies tailored for other data types and tasks, such as graph data~\cite{kose2021fairness} and candidate generation tasks in recommender systems~\cite{zhou2021contrastive}. In graph data, nodes with similar attributes (including sensitive attributes) tend to be connected. As a result, the graph representation learning methods that exploit the topological structures tend to amplify the bias. To reduce this undesirable effect, ~\cite{kose2021fairness} proposes a fairness-aware data augmentation framework, including a feature masking strategy which assigns a larger masking probability to the features that are correlated with the sensitive attributes, and an edge deletion strategy based on whether the sensitive attributes are the same between neighboring nodes. The proposed contrastive learning loss could mitigate the propagation of bias in the graph. Though not targeting sensitive attributes, \cite{zhou2021contrastive} proposes a contrastive learning strategy to mitigate bias against the under-recommended products for candidates generation in large-scale recommender systems, where the goal is to retrieve a small set of entities from a large corpus. They implement a fixed-size first-in-first-out queue to accumulate positive and negative samples for contrastive learning. It is shown that the model trained with the contrastive loss achieves better fairness on the under-recommended products.

\subsection{Hybrid Methods for Implicit Mitigation}
\label{sec:4.3}
\noindent Some studies have investigated hybrid methods that address both disparate impact and disparate treatment. A representative line of work is termed \emph{compositional fairness}, where the model is expected to flexibly accommodate different subsets of sensitive attributes, addressing the fairness issues at the subgroup-level.

One strategy to achieve compositional fairness is disentangled representation learning. As we discussed in Section~\ref{sec:413}, the key idea of disentangled representation learning is to extract sensitive information out of the representation so that predictors trained on the non-sensitive latent factors are \blue{less biased}~\cite{creager2019flexibly}. One advantage of disentangled representation learning is that it can cope with multiple sensitive attributes, and a subset of sensitive attributes can be selected for inference to achieve compositional fairness. Follow-up research has designed more fine-grained disentangled strategies by additionally modeling the intersected information between sensitive and non-sensitive attributes~\cite{park2021learning} and applied the technique to specific applications, such as facial attribute classification for the hospital no-show~\cite{boughorbel2021fairness}.

Adversarial learning is an alternative way to achieve compositional fairness. This strategy has been explored on graph data to learn fair graph embeddings~\cite{bose2019compositional}. The key idea is to train multiple filters to refine the graph embeddings without particular sensitive attributes, where each filter corresponds to one sensitive attribute. For each filter, a discriminator is defined to predict the corresponding sensitive attribute from the filtered node embeddings. The discriminator is trained with an adversarial loss so that the filter is learned to produce \blue{unbiased} representations. This design allows us to flexibly apply combinations of the filters to achieve compositional fairness. Although this paper focuses on graph embedding, the idea can be easily generalized and applied to other data types as well.

\blue{\section{Discussions}\label{sec:5}}

\blue{Prior work has approached machine learning fairness in different directions and proposed various in-processing unfairness mitigation techniques. It is important for the researchers and practitioners to understand the differences among the existing methods. In this section, we provide general discussions on how to choose an appropriate algorithm in real-world applications. We mainly focus on the application scenarios that different methods can tackle and discuss from the perspectives of fairness measurements, task scenarios, base machine learning models, and predictive performance.}

\blue{The selection of fairness measurements is often the first decision we need to consider since it defines the fairness goals. This is particularly important for the explicit methods, where constraints and regularization terms are designed according to the used measurements. However, due to the subjectivity of fairness, there is no universal agreement about which measurement is better. In practice, some fairness measurements can be incompatible. For example, demographic parity demands each sensitive group to have an equal percentage of positive outcomes, while overall accuracy equality requires an equal accuracy. These two measurements will be fundamentally opposed if different sensitive groups have very different positive rates in the ground truth. In this case, if both of these two measurements are used, then the unfairness can never be fully eliminated. Thus, for practitioners, different measurements are often suitable for different scenarios and need to be carefully determined by domain expertise. For machine learning research purposes, one who is only interested in the algorithm design may assume that the suitable measurements are given. Among the papers we surveyed, the most commonly used measurements are demographic parity, equal opportunity, and equalized odds.}

\blue{The task scenarios can often imply which mitigation algorithms are suitable. First, different tasks often have different data characteristics and training schemes. A notable example of the impact of data is graph~\cite{cook2006mining}, which describes complex relationships and interactions among nodes. A recent study shows that graph structure can amplify unfairness through message passing~\cite{jiang2022fmp}. Thus, to achieve fairness in graph data, we need a tailored design to counter such amplification, which is often not considered in the mitigation algorithms built on tabular data. Second, different training schemes will often impact the algorithm design. For instance, semi-supervised and unsupervised settings demand tailored designs to leverage unlabeled data to mitigate unfairness~\cite{zhang2020fairness,schmidt2018fair}. Thus, the algorithm design highly depends on the tasks at hand. Third, due to privacy reasons, we may not have access to sensitive attributes in some applications. If the sensitive attributes are unfortunately not available in a task, then most of the existing mitigation methods will be inapplicable. We instead need to resort to the algorithms that do not require sensitive attributes~\cite{hashimoto2018fairness,lahoti2020fairness,gupta2018proxy}.}

\blue{The base machine learning models, which often vary in different application scenarios, can put constraints on what mitigation algorithms can be adopted. Firstly, implicit methods can only be applied to deep learning models but not traditional machine learning models since it focuses on debiasing the representations. Thus, for traditional machine models, we often need to design explicit constraints or regularization terms, or resort to pre-processing or post-processing solutions. Secondly, many already complex machine learning systems, such as the recommender system deployed at Facebook~\cite{acun2021understanding}, often do not allow one to make significant changes to the model. In this case, some mitigation techniques, such as disentangled representation learning, may require broad changes in the model design and be hard to adopt. We refer interested readers to~\cite{roh2020fairbatch,du2021fairness} for detailed discussions of how to achieve fairness with minimum modifications on the batch sampling~\cite{roh2020fairbatch} and decoders~\cite{du2021fairness}.}

\blue{Finally, fairness will influence the predictive performance of machine learning models. However, in this survey, we will not discuss the quantitative relation between fairness and predictive performance since it is often subject to datasets, algorithms, application domains, and whether the hyperparameters are well-tuned. We believe we can not have an affirmative answer without a comprehensive and fair benchmarking of the existing methods. It is worth noting that the relation between predictive performance and fairness is controversial in the existing work. While most studies found that their relationship is a trade-off~\cite{kamiran2012data,feldman2015certifying,chouldechova2017fair}, some argued that they are sometimes in accord~\cite{wick2019unlocking}. Readers who are interested in fairness and predictive performances and their relationship may refer to~\cite{wick2019unlocking}.}


\section{Research Challenges}
\label{sec:6}

In-processing methods for machine learning fairness have attracted increasing attention in the community and received significant progresses in recent years. However, there are still several remaining research challenges to be investigated.

\textbf{\blue{Fairness} with Missing Sensitive Attributes.} While most of the existing studies assume that the sensitive attributes information is available in the dataset, sensitive attributes may not be disclosed or available in many real-world applications. Some laws have imposed restrictions on the use of sensitive information, e.g., the General Data Protection Regulation (GDPR) explicitly requires businesses to protect the personal data for citizens of European Union. Thus, it is a crucial problem to address fairness issues with 
missing sensitive attributes information. How to achieve fairness in machine learning models remains to be a challenge if very few or even no individual's sensitive attribute is known. Research in this direction is relatively lacking in the literature~\cite{hashimoto2018fairness,lahoti2020fairness,sagawa2019distributionally}. More efforts are needed to investigate how to achieve fairness with missing or limited sensitive attributes information.

\textbf{\blue{Fairness} with Multiple Sensitive Attributes.} The current in-processing techniques mainly focus on addressing the \blue{unfairness} induced by one sensitive attribute. When multiple sensitive attributes exist, the model \blue{that is fair for} one sensitive attribute could still be \blue{unfair for} other sensitive attributes. For instance, both gender and race could lead to unfairness in the \emph{Adult} dataset. The \blue{fair} model that is \blue{designed} solely for women will still be likely to be \blue{unfair for} black women. Additionally, in many real-world scenarios, it is necessary yet challanging to design a  model that can flexibly accommodate different fairness needs. For example, we could only require the model to be fair towards gender while we may need to simultaneously consider gender, race, and age in other scenarios. However, the model trained solely for mitigating gender \blue{unfairness} may not perform well when simultaneously considering gender, race, and age, and vice versa. This practical problem, i.e., how we can accommodate different combinations of sensitive attributes to achieve fairness in different subgroups, is relatively less studied~\cite{kearns2018preventing,kearns2019empirical,bose2019compositional} and needs more future work to explore.

\textbf{Fairness Measurements Selection.} Since the design of the \blue{mitigation} techniques is often inspired by the desired fairness measurements, selecting an appropriate fairness measurement is crucial and depends on the situations at hand. For example, the input of adversarial debiasing~\cite{zhang2018mitigating} depends on the selected fairness measurement because different information is needed for different measurement metrics. Specifically, the prediction $\hat{y}$ is used to calculate demographic parity, whereas both the prediction $\hat{y}$ and the ground truth $y$ are need for equalized odds. It is very likely that a fair model in terms of one measurement \blue{remains unfair} when evaluated with a different measurement. It is a practical need to understand the effect of fairness measurement on algorithm design to achieve fairness with respect to a single measurement or a combination of multiple measurements. This can help the practitioners to achieve the desired fairness outcome in real-world deployments.

\textbf{Balance between Group and Individual Fairness.} Due to the different goals of group and individual fairness, the existing mitigation techniques often only focus on one of them. However, the model that is optimized to achieve group-level fairness may not be fair in terms of individual-level fairness, and vice versus. In some certain application scenarios, it is desirable to achieve both group and individual fairness. For example, when ranking the items for users in recommender systems, we may aim to minimize the discrepancy between the rankings of each pair of individuals with different sensitive attributes while preserving the overall fairness in the group-level. Such goal could be formulated as a constraint optimization problem, e.g., optimizing individual-level fairness under group-level fairness constraints~\cite{garcia2021maxmin}. More studies are encouraged to investigate the relationships between the group- and individual-level fairness and the techniques to achieve both of them.

\textbf{Integration of Interpretation and Fairness.} Despite the success of machine learning models, they are often been criticized to be uninterpretable. Various interpretable machine learning techniques have been proposed to better understand and debug the model~\cite{du2019techniques}, which could serve as the tools to detect and mitigate bias to achieve fairness~\cite{du2020fairness}. For example, in sentiment analysis, local interpretation could detect race biases by getting the feature importance for all the features~\cite{kiritchenko2018examining}. Also, local interpretation could serve as regularization terms to achieve fairness by enforcing the predictions to be less dependent on the sensitive attributes~\cite{ross2017right,liu2019incorporating}. Many interpretation techniques could be potentially adapted to achieve machine learning fairness since the interpretation on features expose the effect of sensitive attributes on the decision-making process. 
Despite the promise in leveraging this relationship to achieve fairness, interpretation itself remains to be a challenge since it may trigger the artifacts without careful design~\cite{du2019techniques}. More work on investigating interpretation and particularly the relationship between interpretation and fairness will facilitate the understanding of fairness and motivate the design of \blue{unfairness mitigation} algorithms.

\textbf{Relationship between Fairness and Model Performance.} Intuitively, if we regard the fairness goals as constraints that limit the decision space of the machine learning model, fairness goals will clearly make the model performance suffer because the resulting set of the possible decision spaces is a subset of the original one. A consensus achieved in most previous work is that the relationship between fairness and model performance is trade-off~\cite{kamiran2012data,feldman2015certifying,chouldechova2017fair}. However, a recent study reveals that fairness and accuracy are sometimes in accord, e.g., in some semi-supervised tasks~\cite{wick2019unlocking}. \blue{Additionally, due to the subjectivity of fairness, different fairness goals may affect or be affected by model performance differently.} As the previous work mainly focuses on empirical studies, a systematic and theoretical investigation of the relationship between fairness and model performance, particularly, under what conditions \blue{(such as which fairness goals)}, improving fairness could also improve model performance, is an important future direction.

\textbf{Datasets and Benchmarks for Machine Learning Fairness.} Benchmark datasets that exhibit \blue{discrimination against} certain groups of people are what propels us forward in designing \blue{unfairness} mitigation algorithms. However, benchmark datasets are lacking in the community. The existing studies mainly perform experiments and analyses on a very limited number of small-scale datasets. More efforts in constructing datasets, especially large-scale datasets with demographic information, are encouraged to enable a full exposure of fairness problems. Efforts on benchmarking the existing algorithms under different fairness measurements will also facilitate the research.

\section{Conclusions}
\label{sec:7}
In this survey, we present an overview of the current progress of in-processing techniques which focus on modeling design for improving machine learning fairness. Specifically, we categorize the existing techniques into explicit and implicit mitigation methods based on where the fairness issues are tackled in the model. Furthermore, we summarize how each category of methods could be used to mitigate disparate impact (group-level bias), disparate treatment (individual bias), and other mixed scenarios. Finally, we introduce the remaining challenges to be addressed for future research efforts, advocating deeper understandings of fairness and the development of datasets and benchmarks to facilitate future methods design.


\section*{Acknowledgements}
This work is in part supported by NSF grants IIS-1939716 and IIS-1900990. The authors would like to thank Dr. Xia Hu and Dr. Mengnan Du for their constructive feedback.

\bibliographystyle{ACM-Reference-Format}
\bibliography{ref}

\end{document}